\documentclass[10pt]{article}
\pdfoutput=1

\usepackage{amsfonts}
\usepackage{amsmath}
\usepackage{amsthm}
\usepackage{mathrsfs}

\newtheorem{definition}{Definition}

\usepackage{biblatex}
\usepackage{graphicx}

\usepackage{hyperref}
\hypersetup{
    colorlinks=false,
    pdfborder={0 0 0},
    pdfborderstyle={/S/U/W 0},
}

\newcommand{\email}[1]{\href{mailto:#1}{#1}}

\newcommand{\orcidinline}[1]{\href{https://orcid.org/#1}{\includegraphics[keepaspectratio,height=10pt]{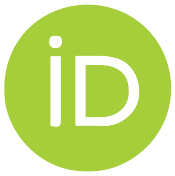}}}

\usepackage[ruled]{algorithm2e} 

\title{Implementing Immune Repertoire Models \\  Using Weighted Finite State Machines}
\author{
    \normalsize Gijs Schr\"oder\orcidinline{0000-0001-6803-3237} \\
    \normalsize\email{gijs.schroeder@ru.nl}
    \and
    \normalsize Inge M.N. Wortel\orcidinline{0000-0003-3362-5229} \\ 
    \normalsize\email{inge.wortel@ru.nl}
    \and
    \normalsize Johannes Textor\orcidinline{0000-0002-0459-9458} \\
    \normalsize\email{johannes.textor@ru.nl}
}
\date{{\footnotesize Radboud University, Institute for Computing and Information Sciences }\\ {\normalsize 7\textsuperscript{th} August, 2023}}

\begin{document}

\maketitle

\begin{abstract}
  The adaptive immune system's T and B cells can be viewed as 
  large populations of simple, diverse classifiers. Artificial immune systems
  (AIS) -- algorithmic models of T or B cell repertoires -- are used
  in both computational biology and natural computing to investigate how
  the immune system adapts to its changing environments. However,
  researchers have struggled to build such systems at scale. For
  string-based AISs, finite state machines (FSMs) can store
  cell repertoires in compressed representations that are orders of
  magnitude smaller than explicitly stored receptor sets. This
  strategy allows AISs with billions of receptors to be generated in a 
  matter of seconds. However, to date, these FSM-based AISs have been
  unable to deal with multiplicity in input data. 
  Here, we show how weighted FSMs can be used to represent
  cell repertoires and model immunological processes
  like negative and positive selection, while also taking into account
  the multiplicity of input data. We use our method to build simple
  immune-inspired classifier systems that solve various toy problems
  in anomaly detection, showing how weights can be crucial for both
  performance and robustness to parameters. Our approach can 
  potentially be extended to
  increase the scale of other population-based machine learning
  algorithms such as learning classifier systems.
\end{abstract}


\section{Introduction}

\subsection{Motivation and prior work}
Artificial neural networks that underlie the rapid advances in AI/ML in recent years, 
from convolutional neural networks to transformers, 
were originally inspired by computational models of the human brain. Intriguingly, our bodies contain a 
second complex adaptive information-processing system that learns through an entirely
different mechanism: the immune system. Unlike the brain, which derives its complexity 
from connections between cells of the same type, the immune system's pattern 
recognition cells -- B and T~cells -- are much more diverse, mainly due to specific 
genetic machinery that randomly generates parts of their DNA sequence. For example, 
the number of different T~cells that can be made due to this mechanism is estimated 
to be in the range of $10^{15}$ to $10^{20}$ \cite{Zarnitsyna2013}, many orders of 
magnitude more than the number of bits stored in the human genome. Each of us harbors
 only around $10^{11}$ T~cells, a small fraction of what can be realized, meaning that many of our T~cells are fairly unique in the population. This large and diverse amount of cells is very expensive to make (the entire process takes until adulthood in humans) and to maintain. Why is such a large number of cells necessary?

Each adaptive immune cell specializes on recognizing very specific molecular patterns 
called \emph{epitopes}. For T~cells, the epitopes are short amino acid sequences (peptides) 
that originate as waste products from every cell's protein degradation machinery. 
Specifically, the CD8$^+$ subpopulation of T~cells recognizes epitopes about 9 amino acids 
long, and the CD4$^+$ subpopulation recognizes longer peptides in the range of 12-14 
amino acids. For any given peptide, the fraction of T~cells in the immune system that will 
recognize it is thought to be on the order of 10$^{-5}$ \cite{Blattman2002}. 
Since immune cells need to protect us from a vast amount of pathogens, some of which may not 
 even exist yet, they need to completely cover the space of possible antigens. 
 This is why so many cells need to be made and maintained. 
 \footnote{This energy expense is even more costly for smaller animals such as mice. 
 A mouse consists of ~$3\times 10^9$ cells and  ~$10^8$ of these -- or 5-10\% of the 
 entire animal -- are adaptive immune cells.}

Computer scientists have long been interested in the computational properties and 
capacities of the immune system, and particularly of its adaptive arm, starting with 
pioneering work by Stephanie Forrest, Alan Perelson, and others in the 
1990s \cite{Forrest1990,Percus1993}. Algorithmic models of the immune system -- 
now called \emph{artificial immune systems} (AIS) -- are often built around the notion 
of large \emph{repertoires} (sets) of small, specific classifiers called \emph{detectors} 
or \emph{patterns}, a property shared with more generic \emph{learning classifier systems} 
(LCS) \cite{Wilson1994}. One of the earliest AISs, Forrest's negative selection algorithm \cite{Forrest1994}, 
was based on the ``education'' of T~cells during negative selection in the thymus: 
upon randomized rearrangement of their receptor sequence, T~cells get shown ``normal'' 
peptides from the host's cells, and those that respond to one of these normal ``self'' 
peptides are killed. This simple process was thought to ensure that T~cells react only 
to the ``nonself'' peptides that would arise, for example, when cells get infected by a 
virus or mutate and become cancerous. More recent data has now cast some doubt on this 
However, the negative selection algorithm still works in principle, and AISs have also 
been used in computational immunology to revisit the negative selection theory in light 
of the new data \cite{Wortel2020t}.

After initial enthusiasm led to the establishment of a research community around 
AIS, it quickly became clear that scale was a central issue in such systems. Like in the 
real immune system, AISs had to provide very large numbers of classifiers to cover the 
entire possible problem space, and the number of classifiers required  often scaled 
exponentially with important parameters of the system. The inability to build AISs at the 
 scale required to solve challenging real-world problems may have been partly 
 responsible for the decline of the AIS field in the 2000s. However, algorithmic models 
 of the immune system continue to be relevant in (computational) immunology itself, 
where they are used to improve our understanding of adaptive immune responses, and  
ultimately to predict such responses. Again, the use of AISs in computational immunology 
requires systems that are large enough to ask and address relevant questions. 

It is perhaps instructive to compare AIS to artificial neural networks, 
which went through an extended period of relative obscurity. Incremental 
improvements in scale and efficiency were made -- until it suddenly became clear 
that all these ``incremental'' developments taken together massively improved the utility 
of such systems. Equally, we feel that AIS models deserve the attention to 
detail needed to build them efficiently and at scale, in order to really 
understand how these systems work and what they are capable of.

An important development in AIS methodology has been the idea that 
\emph{compressed representations} of detector repertoires can be used to build much 
more efficients AISs. This idea was initially used to debunk claims \cite{Timmis2008} 
that detector generation for negative selection with so-called  
r-chunk and r-contiguous detectors is an NP-hard problem \cite{Elberfeld2011}. It was 
later shown how \emph{finite state machines (FSMs)} can be used as a generic building 
block to build such compressed representations for arbitrary string matching rules, 
although for some matching rules such compressed representations will still have an 
exponential size in the worst cases \cite{Liskiewicz2010}. 
Recently, an FSM-based AIS was used to build an algorithmic model of negative selection 
consisting of tens of millions of detectors that processed the entire human proteome 
as input \cite{Wortel2020t}. 

However, these FSM-based AISs still have important limitations. Crucially, they represent 
detector repertoires as sets without multiplicity. This means that all detectors in the 
set will have equal importance or strength, which limits the complexity of the 
information that can be stored.

\subsection{Our contributions}

In this paper, we present a new type of AIS repertoire models where each detector in a 
repertoire is assigned a weight, similar to how classifiers in LCS can be weighted. 
Specifically, we: 

\begin{enumerate}
\item Define weighted versions of positive and negative selection, two of the most important AIS algorithms  (Section~\ref{sectionbackground});
\item Show how weighted positive and negative selection can be implemented efficiently by using compressed repertoire representations based on weighted FSMs (Section~\ref{sectionfsms});
\item Illustrate the potential benefit of using weights by comparing the performance of weighted versus unweighted repertoire models on simple string-based anomaly detection problems (Section~\ref{sectionempirical}).
\end{enumerate}

We have implemented our WFSM-based repertoire models as a series of C++ classes with Python bindings that make heavy use of the library OpenFST \cite{openfst}.
Upon acceptance of this manuscript, we will publicly release our code as open source software.

\section{Definitions}

\label{sectionbackground}

We consider strings $x$ over some finite alphabet $\Sigma$. Throughout, we denote strings using lowercase latin letters. E.g., $x=0010 \in \{0,1\}^4$ is a string consisting of 4 binary characters and $x_3=1$ is its third character. For a set $S$, we use $|S|$ to denote its cardinality.

\subsection{Unweighted repertoire models}

\label{sectionunweightedrep}

We use the term \emph{repertoire model} in this paper to denote a type of AIS that is 
loosely based on T~cell and B~cell receptor repertoires in the real immune system. 
Generally speaking, such models consist of large populations of detectors 
(also called classifiers or patterns), where each detector recognizes a small part of 
some universe $\mathcal{U}$. By extension, sets of detectors can therefore represent 
(``cover'') subsets of $\mathcal{U}$. Although this framework is general and allows 
$\mathcal{U}$ to be any set, in this paper we focus on strings over some alphabet 
$\Sigma$. For simplicity, we will also assume that the strings have a fixed length 
($\mathcal{U}=\Sigma^\ell$), since the generalization of our framework to variable-length 
strings is reasonably straightforward. 

In addition to the \emph{universe}: $\mathcal{U}$ that represents the problem space 
(e.g., objects to be classified), we define a set of \emph{detectors} $\mathcal{D}$; possibly $\mathcal{U}=\mathcal{D}$. A \emph{matching function} $m : \mathcal{D} \mapsto 2^\mathcal{U}$, where $2^\mathcal{U}$ denotes the powerset of $\mathcal{U}$, associates every detector with the elements it recognizes. In a slight abuse of notation, we define the inverse matching function $m^{-1} : \mathcal{U} \mapsto 2^\mathcal{D}$ as $m^{-1}(x) = \{ d \in \mathcal{D} \mid x \in m(d) \} $. A \emph{repertoire} is simply a subset of $\mathcal{D}$. 

\begin{definition}[Matching rules]
Given an alphabet $\Sigma$ and a string length $\ell$, we define the following matching rules:
\begin{enumerate}
\item \emph{Wildcard pattern matching:} considers a wildcard symbol $\# \notin \Sigma$ 
and sets $\mathcal{U}=\Sigma^\ell, \mathcal{D}=\Sigma \cup \{\#\}^\ell$. 
Then $m(x) = \{ y \in \mathcal{U} : \forall i \in \{1,\ldots,n\} : x_i \in \{ y_i, \# \} \}$. 
 \item \emph{r-Contiguous matching:}  $\mathcal{U}=\mathcal{D}=\Sigma^\ell$, $m_r(x)=\{y \in \mathcal{U} : \exists i \in \{1,\ldots,\ell-r+1\} : \forall j \in \{i,\ldots,i+r-1\} : x_j = y_j \}$. 
 The parameter $r$, called \emph{matching radius}, controls the number of strings each 
 detector matches: increasing $r$ means matching fewer strings. This pattern matching 
 rule is common in AIS.
\item \emph{r-Hamming matching:} $m_r(x)=\{y \in \mathcal{U} : |\{ i : x_i \neq y_i \}| \leq r \}$. 
Here, increasing the matching radius $r$ means matching more strings.
\end{enumerate}
\label{definitionmatchingrules}
\end{definition}

Using any such matching rule, we can define classifiers based on a repertoire model. 
Mirroring the stages of T~cell selection in the thymus, there are two main 
repertoire-based classification algorithms that were studied in the AIS field. 
These are so-called \emph{one-class} classification algorithms that take an input 
sequence $S \in \mathcal{U}^*$ (also called \emph{self}) to construct a detector set $D$, which is then 
used to determine whether or not the elements of a second input sequence $T \in \mathcal{U}^*$ 
belong to the same class as the elements of $S$:

\begin{definition}[Positive and negative selection]
Given an input sequence  $S=(s_1,\ldots,s_n) \in \mathcal{U}^n$, 
a detector type $\mathcal{D}$ and a matching function $m$,  we call a 
repertoire $D \subseteq \mathcal{D}$
\begin{enumerate}
\item \emph{positively selected} if $D \subseteq \bigcup_i m^{-1}(s_i)$ is a set of detectors 
that match at least one input string.
\item \emph{negatively selected} if $D \subseteq \mathcal{D} \setminus \bigcup_i m^{-1}(s_i)$ 
is a set of detectors that do not match any input string. 
\end{enumerate}

A negatively (positively) selected repertoire $D$ is \emph{maximal} if there is no strict superset of $D$ that is also negatively (positively) selected with respect to the same input $S$. Given a repertoire $D$ and an element $t \in \mathcal{U}$, we define the \emph{scoring function} $D(t) =  |D \cap m^{-1}(t)|$ as the number of detectors in $D$ that match $t$. 
\label{definitionposnegsel}
\end{definition}

For a positively selected repertoire, the scoring function can be understood as a \emph{normalcy score} (a high value $D(t)$ means that $t$ is ``similar to'' $S$) whereas for negative selection, the interpretation is the opposite (\emph{anomaly score}). The scores output by a positively or negatively selected repertoire can be used for threshold-based classification. Note that we did not define which specific detector set is used, only which detectors \emph{could} be in the set. In practice, there are only two methods that have been reasonably well explored. The first and perhaps simplest is generating detectors by rejection sampling. However, depending on the input, rejection rates can be high \cite{Dhaeseleer1996,Dhaeseleer1996b}. 
The second method is to use maximal detector sets \cite{Textor2012,Textor2014}. This requires the use of compressed repertoire representations (see next Section) for anything but the most trivial inputs. Interestingly, although positive and negative selection using randomly generated detectors can give different results, for many matching rules this is not the case when maximal detector sets are used.

\newtheorem{remark}{Remark}

\begin{remark}
For the matching rules in Definition~\ref{definitionmatchingrules}, 
positive and negative selection with maximal detector sets are equivalent classifiers.
\label{remarkbotharethesame}
\end{remark} 

\begin{proof}
If $D^+$ and $D^-$ are the maximal positively and negatively selected detector sets w.r.t. an input $S$, 
then we have $D^+ \cup D^- = \mathcal{D}$ and therefore 
$|D^+ \cap m^{-1}(t)|
+ |D^- \cap m^{-1}(t)|=|m^{-1}(t)|$. Since $|m^{-1}(t)|$ is the same value for all $t \in \mathcal{U}$ 
for the matching rules in Definition~\ref{definitionmatchingrules}, the scoring function of negative 
selection is a constant minus the scoring function of positive selection, and vice versa.
\end{proof}

Beyond positive and negative selection, further algorithms that can be performed with repertoire models include sampling from $D$ or incrementally modifying $D$ \cite{Textor2014}; these tasks are not substantially more complicated and can be implemented using very similar algorithmic techniques, which is why we don't discuss them in detail here. Earlier versions of these algorithms only distinguished between empty and nonempty $D \cap m^{-1}(t)$ in the classification step -- in effect binarizing the scoring function at a fixed threshold of 1. While compared to these earlier versions, the use of scores is already a substantial improvement (as also in the real immune system, immune responses are likely not raised by single cells), these algorithms are still limited in that they do not take possible multiplicity of strings in the input sequences into account. We therefore suggest the following extension of these models.

\subsection{Weighted repertoire models}

\label{sectionweightedrep}

A \emph{weighted repertoire} $(D,w), D \subset \mathcal{D}, w : D \mapsto \mathbb{R} \setminus \{0\}$ is a set of strings with associated nonzero real-valued weights. The weights can have different interpretations. For instance, positive integer weights could be used to simply store multiplicity of detectors, whereas real-valued weights could represent each detector's relative importance, and weights in the interval $(0,1]$ might represent probabilities that each detector is actually present in the set. 
 
In this paper, we will use the following definitions.

\begin{definition}[Weighted positive and negative selection]

Given an input sequence  $S=(s_1,\ldots,s_n) \in \mathcal{U}^n$, a detector type $\mathcal{D}$ and a matching function $m$, we define the following:
\begin{enumerate}
\item \emph{Weighted positive selection} uses the detector repertoire $D = \bigcup_i m^{-1}(s_i)$ with weights $w(d) = |\{i : d \in m^{-1}(s_i)\}|$ where each detector is weighted by the number of input samples it recognizes. 
\item \emph{Weighted negative selection} uses the detector repertoire $D = \mathcal{D} \setminus \bigcup_i m^{-1}(s_i)$ with a pre-existing weighting function $w : \mathcal{D} \mapsto \mathbb{R}$. 
\end{enumerate}

The \emph{scoring function} $D_w(t)$ assigns normalcy or anomaly scores $\sum_{d \in D \cap m^{-1}(t)} w(d)$
to every $t \in \mathcal{U}$.

\label{definitionweightedposnegsel}

\end{definition}

Introducing weights breaks the simple duality between positive and negative selection algorithms, and even using maximal detector sets, the results will no longer be equivalent. This is because weights in positive selection encode information about multiplicity in the input sample, whereas weights in negative selection encode a pre-existing bias among detectors. Such a bias might seem like an odd choice, but it is relevant in computational immunology because it is well-known that different immune cell receptor sequences have very different chances to be generated \cite{Elhanati2014}. In addition to positive and negative selection, weighted versions of other epertoire modeling tasks \cite{Textor2014} could be considered  -- e.g., sampling could be performed proportionally or inversely proportionally to the weights.

\section{Repertoire models based on weighted finite state machines}

\label{sectionfsms}

\begin{figure}
  \centering
    \includegraphics{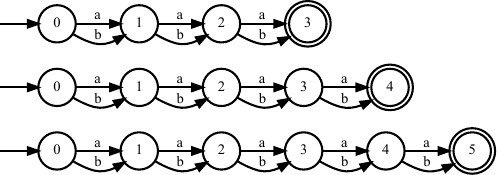}
  \caption{
	FSMs encoding the string set $\{a,b\}^\ell$ for $\ell \in \{3,4,5\}$. Although the string sets
	grow exponentially in size as a function of $\ell$, FSM size is linear in $\ell$.
  }
  \label{fig:fsa-ex}
\end{figure}

In this paper we extend earlier work that showed that repertoire models can be implemented efficiently using 
finite state machines (FSMs), leading to improvements in space and time complexity of several orders of magnitude
\cite{Elberfeld2011,Textor2014}. The basic idea is that FSMs can compactly represent large sets of strings with 
shared structure (Figure~\ref{fig:fsa-ex}). The required operations necessary for repertoire modeling, such as 
computing unions, intersections, and set differences of detector sets, can all be efficiently performed directly
on FSMs. Since FSMs are generic data structures, this approach benefits from a large body of work done to 
develop efficient FSM algorithms and high-quality, mature software implementations such as the OpenFST 
framework \cite{openfst}. We extend the FSM-based repertoire modeling approach to weighted repertoires as defined
in Section~\ref{sectionweightedrep}. This requires us to deal not only with strings $D\subset\mathcal{D}$,
but also the associated weights $w:D\mapsto\mathbb{R}$. For this purpose, we use \emph{weighted finite state machines}
(WFSMs). 

\subsection{Weighted finite state machines}\label{ssec:wa-ops}

For completeness, we give a brief definition of WFSMs used in this paper, but note that this
is equivalent to a ``standard'' definition found in textbooks (up to some simplifications, like 
an absence of final state weights).

\begin{definition}[Weighted finite state machine (WFSM)]
\hfill A WFSM \\ $M=(Q,E,c,w,q_s, Q_F)$ over an alphabet $\Sigma$ and a semiring $\mathbb{K}$ consists of a 
directed graph $G=(Q,E)$ linking states $q \in Q$ where the state $q_s \in Q$ is called the 
\emph{initial (starting) state} and the states $Q_F \subseteq Q$ are called the 
\emph{accepting (final) states}. 
The edges \emph{(transitions)} $q_i \to q_j \in E$  are labeled by characters $c : E \mapsto \Sigma$ and weights $w : E \mapsto \mathbb{K}$. 
The \emph{path weight} $w(p)$ of a path $\pi = q_0 \to q_1 \to \ldots \to q_k$ is defined 
as $\prod_{j=0}^{k-1} w( q_j \to q_{j+1} )$, where $\prod$ uses the product operator in $\mathbb{K}$,
if $q_0=q_s$ and $q_k \in Q_F$; if $q_0 \neq q_s$ or $q_k \notin Q_F$, then the path weight $w(p)=0$.
The \emph{weight} $w_M(s)$ of a string $s \in \Sigma^*$ is the sum of all weights of paths labeled with 
$s$ in $M$. 
\end{definition}

\begin{figure}
  \centering
    \includegraphics{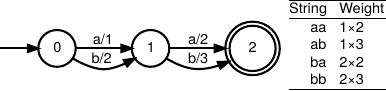}
  \caption{
    A weighted FSM, along with a table showing the strings it contains and the associated weights for
    these strings. Path transitions are labeled (character/weight). 
    Each unique path represents a string (the concatenation of every such character along 
    the path) with its associated weight (the product of every weight along the path).
  }
  \label{fig:wa-ex}
\end{figure}

An example WFSM over $\Sigma=\{a,b\}$ and $\mathbb{K}=\mathbb{R}$ is shown in Figure~\ref{fig:wa-ex}. While  
we typically use real-valued weights, most WFSM theory and algorithms is general and applies for any semiring.
Similar to ordinary FSMs, WFSMs support set operations.
In contrast to FSMs, these set operations also need to perform arithmetic on the weights. 
This allows us to manipulate the weights for many strings at once. In this paper, we restrict our attention to
\emph{acyclic} WFSMS (containing only finite-length strings) and furthermore our WFSMs are 
\emph{deterministic}, meaning there is only one path of nonzero weight associated with every string
$s \in \Sigma^*$. However, determinism is not an essential requirement for us, and in fact software implementations
such as OpenFST might generate non-deterministic WFSMS in intermediate computation steps (such as when performing 
union). Every WFSM is associated with a \emph{language} $L(M)=\{ s \in \Sigma^*: w_M(s) \neq 0 \}$.

To perform positive and negative selection (Definition~\ref{definitionweightedposnegsel}) using WFSMs,
we will need to perform the following operations: (1) weighted union $\cup$; (2) weighted intersection $\cap$; 
(3) weighted set difference $\ominus$; (4) weight summation $\mid \cdot \mid$. Task (4) is straightforward and easily
solved using standard graph algorithms. Below we briefly define the canonical weighted versions of set 
union, intersection, and difference. Most straightforward of these is perhaps the definition of
weighted union:
\begin{definition}[Weighted union]
  A weighted union  of two WFSMs $M_1$ and $M_2$ is defined as an FSM $M_1 \cup M_2$ such that 
  $$\forall s \in \Sigma^*: w_{M_1 \cup M_2}(s) = w_{M_1}(s) \oplus w_{M_2}(s) \ , $$
  where $\oplus$ denotes addition in $\mathbb{K}$.
\end{definition}

Note that this definition preserves the set theoretic interpretation of union with respect to a string being in the language: if $s$ is not in the language of both $M_1$ and $M_2$, it remains absent in the union, but if it is present in $M_1$, $M_2$ or both, then it is in the union -- unless $w_{M_1}(s) = -w_{M_2}(s)$.
Unlike for sets, the weighted union however does not have a unique result. Typically WFSMs are minimized
upon union computation, which does deliver a unique result (assuming weights are canonically distributed 
over paths, see next section).

\begin{definition}[Weighted intersection]
  A weighted intersection of two WFSMs $M_1$ and $M_2$ is defined as an FSM $M_1 \cap M_2$ such that 
  $$\forall s \in \Sigma^*: w_{M_1 \cap M_2}(s) = w_{M_1}(s) \otimes w_{M_2}(s)\ , $$
  where $\otimes$ denotes multiplication in $\mathbb{K}$.
\end{definition}
This definition of weighted intersection recovers unweighted intersection if the ring $\mathbb{Z}_2$ is used
for weights. The least intuitive operation of the three is perhaps the following.
\begin{definition}[Weighted set difference]
  The weighted set difference of two WFSMs $M_1$ and $M_2$ is defined as an FSM $M_1 \ominus M_2$ such that 
  $$\forall s \in \Sigma^*: w_{M_1 \ominus M_2}(s) = \begin{cases}0 & \text{if } s \in L(M_2) \\ w_{M_1}(s) & \text{otherwise}\end{cases}$$
\end{definition}
Hence, the weights of strings not present in $M_2$ will be preserved when subtracting $M_2$ from $M_1$, but all strings also present in $M_2$ are removed from $M_1$ regardless of weights. We use o-minus ($\ominus$) here to emphasize that this operation does not behave similarly to point-wise subtraction of reals.

\subsection{Positive and negative selection using weighted finite state machines}

Having defined the operations necessary to perform weighted versions of standard set operations,
we can now implement weighted positive and negative selection (Definition~\ref{definitionweightedposnegsel})
in a very similar manner as this has previously been done for the unweighted versions of these 
algorithms (Definition~\ref{definitionposnegsel}). An important prerequisite is that 
for each input string $s \in \mathcal{U}$, we are able to generate a WFSM
$M[s]$ containing all detectors that recognize $s$ with unit weights, i.e., 
$$
w_{M[s]}(d) = \begin{cases} 1 & d \in m^{-1}(s) \\ 0  & \text{otherwise}\end{cases}
$$

See earlier work for examples on how to construct such FSMs for common matching rules \cite{Textor2014}; we can 
simply add unit weights to all edges to obtain the required WFSM. 
Then we can implement weighted positive selection as shown in Algorithm~\ref{algorithmweightedpossel}.

\begin{algorithm}
\caption{Weighted positive selection}
\SetKwProg{generate}{Function \emph{generate}}{}{end}
\SetKwInOut{Input}{input}\SetKwInOut{Output}{output}

\Input{Samples $S=(s_1,\ldots,s_k) \in \mathcal{U}^k, T=(t_1,\ldots,t_l)  \in \mathcal{U}^l$}

\Output{Scores $D_w(t)$ for each $t_i \in T$}

$M \gets \bigcup_{i=1}^{n} M[s_i]$ 

\ForEach{$i \in \{1,\ldots,l\}$}{
	output $|M \cap M[t_i]|$
}
\label{algorithmweightedpossel}
\end{algorithm}

Weighted negative selection is implemented in a very similar manner, requiring only a single additional operation as shown in 
Algorithm~\ref{algorithmweightednegsel}. This algorithm uses an operator $M[\mathcal{D}]$ denoting construction of a WFSM containing all possible detectors. This can be constructed using unit weights, or possibly other weights representing pre-existing biases in the repertoire, such as those arising from biased sequence recombination events in the real immune system \cite{Elhanati2014}.

\begin{algorithm}
\caption{Weighted negative selection}
\SetKwProg{generate}{Function \emph{generate}}{}{end}
\SetKwInOut{Input}{input}\SetKwInOut{Output}{output}

\Input{Samples $S=(s_1,\ldots,s_k) \in \mathcal{U}^k, T=(t_1,\ldots,t_l)  \in \mathcal{U}^l$}

\Output{Scores $D_w(t)$ for each $t_i \in T$}

$M \gets \bigcup_{i=1}^{n} M[s_i]$ 

$M \gets M[\mathcal{D}] \ominus M$

\ForEach{$i \in \{1,\ldots,l\}$}{
	output $|M \cap M[t_i]|$
}
\label{algorithmweightednegsel}
\end{algorithm}

\subsection{Implementing weights using exact rational algebra }

The same set of strings -- weighted or not -- can be represented by FSMs in more than one way.
Some of these representations have fewer states than others, where (given a set of 
strings), those
representations with the fewest possible number of states are called \emph{minimal}.

When performing operations on WFSMs in OpenFST,
the WFSMs resulting from these operations will generally not be minimal.
Weighted union, in particular, often produces WFSMs that are larger than necessary.
This becomes especially problematic when performing weighted union repeatedly,
like in Algorithms \ref{algorithmweightedpossel} and \ref{algorithmweightednegsel}.
In order to keep the size of the WFSMs manageable, they need to be \emph{minimized} after 
such weighted union steps (i.e., transformed to their minimal equivalent).
To do so requires an equivalence relation for transitions 
through (W)FSMs. For regular FSMs, this equivalence is simply decided by whether the 
transitions have the same label and the same destination state, but for WFSMs, 
equivalence additionally requires the transitions to have the same weight.

\begin{figure}
  \centering
    \includegraphics{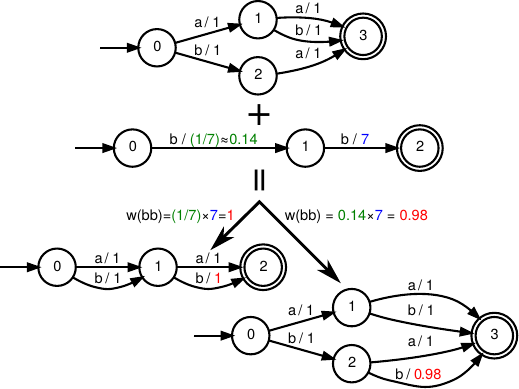}
  \caption{
    Exact arithmetic can reduce WFSM size.
    We consider the union of two WFSMs with exact and inexact representation of weights.
    With exact arithmetic,
    the path for $bb$ contains the weights $\frac{1}{7}$ and $7$,
    resulting in the path weight $w(bb)=\frac{1}{7}\times7=1$.
    Since the union of the two WFSMs contains the strings $\{a,b\}^2$ with weight 1 each,
    it can be minimized to a small WFSM (bottom left).
    However, in the inexact representation (decimal with two significant digits), 
    the path weight becomes $w(bb)=0.14\times 7=0.98$, which cannot be merged with the
    existing path of weight 1, generating one extra state and two extra transitions.
  }
  \label{fig:fp-minimization}
\end{figure}

Importantly, this extra equivalence requirement for WFSMs becomes harder to meet when 
working with inexact float arithmetic. Given a path through the WFSM, the same weight
might be distributed along the path in different ways -- for example, $(\frac{1}{7},7)$ 
or $(\frac{355}{113},\frac{113}{355})$, since 
$\frac{1}{7}\times 7 = \frac{355}{113}\times\frac{113}{335}=1$.
Yet these path weights, although equivalent, may no longer be recognized as such 
due to small errors when representing these fractions with floats.
These inequalities can prevent WFSMs from minimizing to their smallest possible state,
as demonstrated in Figure~\ref{fig:fp-minimization}.

Typical solutions for this problem with floats is to quantize them or to test for 
approximate equality rather than exact equality, both of which are applied in OpenFST.
However, these strategies are ultimately unsuccessful at preventing error accumulation
when many repeated WFSM operations are performed, as necessary in 
Algorithms~\ref{algorithmweightedpossel} and~\ref{algorithmweightednegsel}.
This is no minor issue; 
the accumulation of inaccuracies led to an explosion of WFSM size that prevented us 
from doing any testing on real data.

\begin{figure}
  \centering
	\includegraphics{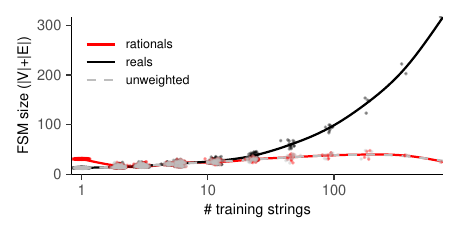}
  \caption{
	Exact rational arithmetic can prevent state explosion when merging many 
	WFSMs. A WFSM representing all strings in $\{0,1,2\}^6$ is constructed by 
	successively merging single-string FSMs in a binary tree recursion. With unweighted
	FSMs, shared substrings are exploited at every stage yielding a strongly compressed FSM. 
	With real-weighted FSMs, even though all weights are 
	initially the same, inaccuracies accumulate and prevent states 
	from being merged (black line). Using exact rational arithmetic prevents this issue
	and rescues the desired behavior, leading to the same final result as in the 
	unweighted case (red line). Points show FSM sizes for different numbers of training strings
	(with some jitter added in the x-direction to avoid overplotting).
	Lines show a Loess smoothing of the data.
  }
  \label{figurefpvsrationals}
\end{figure}

To illustrate this issue, consider 
the problem of computing $M=\sum_i M_i$, where each $M_i$ contains one string 
in $S=\{0,1,2\}^6$ with unit weight.
The different $M_i$ are all disjoint and their union is all of $S$.
Therefore, $w_M(s) = 1$ for all $s\in S$, making this essentially an unweighted FSM
with a minimal $M$ of 7 states and 18 transitions.
In Figure~\ref{figurefpvsrationals}, we show how WFSM size grows with the number of 
strings contained in intermediate stages of computing $M$. An unweighted FSM implementation
indeed contains 7 states and 18 transitions, whereas a WFSM implemented with float weights
produces a ``minimal'' $M$ with 80 states and 237 transitions.

To address these issues, we implementeded weights using an exact rational representation 
from Boost \cite{Gurtovoy2002}, which restored the equivalence with the unweighted 
FSM in the final output, and where intermediate stages were at most four times larger.

\section{Empirical analysis}

\label{sectionempirical}

Standard non-weighted repertoire models were based on the assumption that the universe
$\mathcal{U}$ of elements to be classified is 
disjointly partitioned into ``self'' and ``nonself'' subsets, and the
goal was to estimate the boundary between these two classes \cite{Forrest1994,Timmis2008}.
In theory, such problems be perfectly solved without taking the multiplicity of input
strings into account: a single witness string is enough to perfectly decide
membership.

Unfortunately, many real-world problems do not fit this assumption.
For example, suppose we were to distinguish language based on n-grams (short sequences of letters).
It is known that even short n-grams such as 3-grams contain enough information
to solve this task satisfactorily \cite{Dunning1996}. However, given enough input text, almost 
every combination of 3 input letters will likely occur at least once in the input
regardless of the language (it has been pointed out that \emph{llj}
is not a typical letter sequence in English but ``only a \emph{killjoy}
would claim'' it never occurs \cite{Dunning1996}). This should make it critical to not only consider
the presence or absence of a string, but also its frequency. Interestingly, 
previous research has shown that negative selection algorithms can nevertheless 
solve such problems reasonably well \cite{Wortel2020t}. 
However, the amount of input text used in these
studies was relatively small. 

One way to model the aforementioned type of classification problems is by considering
a ``fuzzy membership function'' $f: x \mapsto [0,1]$ that assigns a \emph{degree of membership}
of each string to every class. Despite existing results on the negative selection
problem, we hypothesized that unweighted AIS should perform poorly on such fuzzy problems.

\subsection{The noisy bitstring problem}

To test our hypothesis, we first defined a very simple toy example of a
fuzzy classification problem. In this \emph{noisy bitstring problem}, we consider
random bitstrings $X(c,\mu)$ where $c \in \{0,1\}^\ell$ and $0 \leq \mu \leq 1$. 
$X(c,\mu)$ is generated by the following algorithm: draw a random number
$x$ from a geometric distribution with parameter $1-\mu$. Let $x' = \min(x,\ell)$.
Flip $x'$ randomly chosen bits of $c$ and return the result.
In particular, $X(c,0)$ is always $c$, and $X(c,1)$ is always the bitwise
complement of $c$. 

We can set up a fuzzy classification problem by defining the following 
membership functions $f_0$ and $f_1$:
$$
f_0( x ) = \text{Prob}( X(0^\ell,\mu)=x ) ; f_1( x ) = \text{Prob}( X(1^\ell,\mu)=x ) 
$$
Particularly, for $0 < \mu < 1$, every bitstring has a nonzero 
probability of occurrence in both $X(c,0)$ and $X(c,1)$, but for $\mu \ll 1$, we
have for example that $f_0( 0^\ell ) \gg f_1( 0^\ell )$. Our repertoire $D$ 
is now tasked  with assigning a score $D(t)$ to every string such that the 
distributions $D(X(0^\ell,\mu))$ and $D(X(1^\ell,\mu))$ are as different as 
possible -- we will use the AUC metric to measure this difference.

 To solve this problem, it 
should be critical to take multiplicity into account.

\begin{figure}[h]
  \centering

   \includegraphics[width=0.7\textwidth]{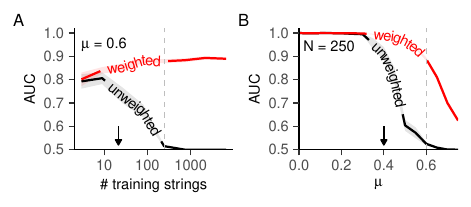}
   \caption{
   Positive selection poorly classifies noisy bitstrings for large
   inputs, but can be rescued by adding weights. A:
   AUC (mean $\pm$ SEM) of 20 independent runs, using $\ell=8, \mu=0.6$ and $5$-contiguous matching,
   varying input sizes N, and a test set of 100 random samples per class. 
   B: likewise, but for a fixed training set size of 250 strings, 
   now varying the mutation rate $\mu$. Dashed lines represent the same parameter 
   combination with $N=250, \mu=0.6$. Arrows represent the values of $\mu$ and $N$ for
   which the expected number of strings with a majority of foreign bits in a training
   sample exceeds 1. 
  }
  \label{figurestringdistinction}
\end{figure}

When simulating unweighted versions of positive selection, we found the seemingly paradoxical 
effect that performance was reasonable for small input samples, but then rapidly degraded for larger
input samples (Figure~\ref{figurestringdistinction}A, black line). This occurred because with larger samples it became more and more likely to find the 
center string of the ``foreign'' class in the input.
By contrast, weighted positive selection 
should not be ``fooled'' by such rare events because it can also learn from the frequencies of
patterns in the input strings. Indeed, while multiplicity in the training set is rare for small
samples and the two versions initially behaved very similarly, the performance
of weighted selection kept improving with larger samples as expected (Figure~\ref{figurestringdistinction}A, red line).
Thus, we can indeed conclude that fuzzy classification problems can be difficult to solve using
unweighted repertoire models, especially at large sample sizes.
Weighted positive selection was also more robust to higher mutation 
rates $\mu$ (Figure~\ref{figurestringdistinction}B), which -- similar to larger input 
sizes -- endanger performance by increasing the frequency of foreign-looking 
strings in the training input.

\begin{figure}[h]
  \centering

   \includegraphics[width=0.7\textwidth]{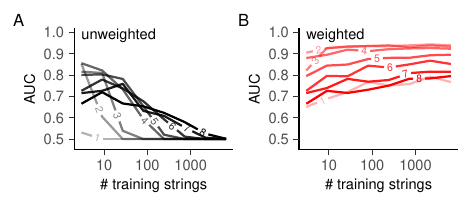}
   \caption{
   Weighted positive selection (A) is less sensitive to the threshold parameter $t$
   than unweighted positive selection (B). 
   AUCs (mean $\pm$ SEM) are shown for 20 independent runs, using $n=8, \mu=0.6$ and contiguous matching
   with varying thresholds $t$ (colored lines). Training and test sets were generated
   as in Figure~\ref{figurestringdistinction}.
  }
  \label{figuresensitivitytot}
\end{figure}

A well-known problem with repertoire models is their sometimes extreme sensitivity to the 
threshold parameter $t$ that determines when strings appear as ``similar'' to the 
receptor repertoire \cite{Dhaeseleer1996,Dhaeseleer1996b,Stibor2005,Wortel2020t}.
We also found this effect for 
unweighted positive selection on noisy bitstrings, where the ``best'' $t$ additionally
depended on input size (Figure~\ref{figuresensitivitytot}A). Interestingly, we 
found that weighted positive selection was more robust to the choice of $t$, with 
$t = 2,3,4$ now giving very similar performances throughout a range of input sizes (Figure~\ref{figuresensitivitytot}B).

These results suggest that on fuzzy classification problems, our weighted 
WFSM-AISs outperform their unweighted counterparts and are less sensitive
to parameters like input size and detection threshold. The question remains: was this
an extreme example, or does the same apply to real-world datasets?

\subsection{Language anomaly detection}

We therefore revisited the problem of language anomaly detection as 
considered previously \cite{Wortel2020t}. In that study, repertoires 
selected on English strings could detect test strings from ``anomalous`` 
languages among English strings reasonably well. However, the training sets used were relatively 
small ($< 1000$ English strings, using contiguous matching with $t=3$) -- small enough 
that foreign-looking 3-grams are unlikely to appear in the training data. We therefore 
asked: would the performance 
of such an (unweighted) AIS degrade as ``unlikely'' letter patterns \emph{do} start to appear
among English training strings?

\begin{figure}[h]
  \centering
   \includegraphics[width=0.7\textwidth]{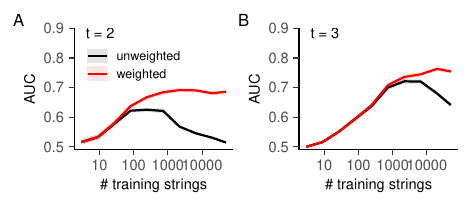}
   \caption{
   Language anomaly detection performance of a positive selection algorithm drops for 
   large training sets and is rescued by adding weights. 3-grams were extracted from 
   Latin or English strings, and (un)weighted positive selection performed  with 
   $t$-contiguous matching against English strings for either $t=2$ (A) or $t=3$ (B). 
   Plots show the AUC (mean $\pm$ SEM) of 20 independent runs.
  }
  \label{figurelang1}
\end{figure}

To test this hypothesis, we downloaded the published set of strings from \cite{Wortel2020t}, 
as well as $\sim$800,000 English strings from the King James bible for training.
From these data, we extracted 3-letter strings, and used our WFSM-AIS to perform both weighted 
and unweighted positive selection on randomly sampled inputs of up to 50,000 English 
training strings. When detecting Latin among English strings, we found that weighted 
positive selection once again started to outperform
its unweighted counterpart at large input sizes (Figure~\ref{figurelang1}). 

\begin{figure}[h]
  \centering
   \includegraphics[width=0.7\textwidth]{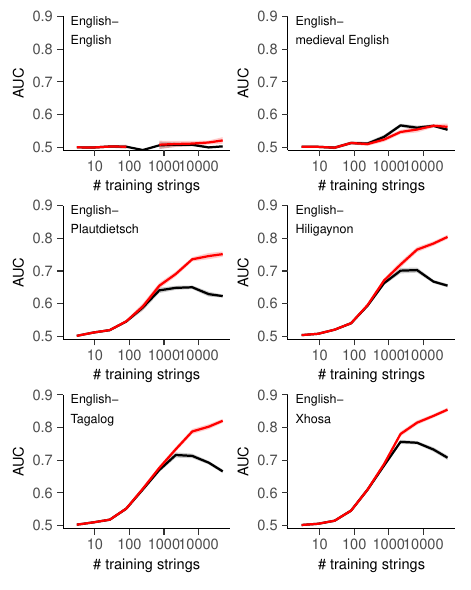}
   \caption{ Analysis of Figure~\ref{figurelang1} performed for different languages. 
	English-English is a negative control experiment where both ``normal'' and ``anomalous''
	test strings are taken from the same language. Plots show the average AUC of 
	20 independent runs with $t = 3$, comparable to Figure~\ref{figurelang1}B.
  }
  \label{figurelang2}
\end{figure}

The input size where unweighted selection starts to perform badly depended on the 
threshold $t$; unlikely 2-grams might appear even in 
relatively small training sets of $\sim$100 strings, whereas the most unlikely 3-grams 
are rare enough that they do not appear in training sets of up to $\sim$1000 strings.
Nevertheless, even these rare patterns eventually caused the performances of the 
weighted and unweighted AISs to 
diverge as inputs reach a size of several thousands of strings (Figure~\ref{figurelang1}). 
Similar results were observed when substituting different languages for the 
anomalous strings (Figure~\ref{figurelang2}). The size of the effect depended on the 
general similarity between English and the ``anomalous'' language considered -- in line
with the intuition that adding weights should not help learn a difference that is not 
there (e.g., in the case comparing English versus more English or 
medieval English).

All in all, our results demonstrate that when anomalous inputs have a non-zero probability
to appear among training data, unweighted AISs perform poorly as the training set size 
increases. By contrast, weighted AISs are able to circumvent this issue by learning 
the information contained in the multiplicity in the training data. These findings 
suggest that weights will be crucial when applying AISs to real-world datasets.

\section{Future work}

Our experiments with WFSM-AISs revealed two key issues which we feel will be important 
to address in future work. First, building a WFSM-AIS in our current framework requires 
merging very large numbers of small classifiers. The OpenFST framework we currently use
implements WFSM union in a way that does not yield a minimal result, such that we need
to call minimization after every union to keep FSM size manageable.
For unweighted FSMs with a ``levelled'' structure that repertoire models typically use
(i.e., acyclic FSMs where all paths between two nodes have the same length),
Textor \emph{et al.} \cite{Textor2014} implemented a custom FSM union algorithm that directly 
outputs a minimal FSM. This could be extended to WFSMs in future work. 
Likewise, we could use a directly 
determinizing union algorithm like the one by Mohri \cite{Mohri1996}, created for a 
similar use-case. Minimization may also be sped up by the algorithm by Eisner \cite{Eisner2003}.

Second, handling weights in WFSMs proved to be more challenging than one might perhaps
anticipate. Our use of exact rational algebra proved critical to get our WFSM-AISs 
to work even on small  input samples -- without it, even repertoires trained on as few as 1000 input characters 
could rapidly blow up to 100s of megabytes in size. While the use of rationals greatly 
improved this and allowed us to perform the experiments reported in this paper, it is 
not a complete solution because the rationals themselves may ``blow up'' and use 
numerators or denominators that are too large to be represented as integers. In such 
cases quantization becomes necessary, bringing back the issue that strings with 
equal weights may no longer be recognized as such. While this did not cause any major 
issues in the experiments reported in this paper, we expect it to become problematic 
when storing large numbers of strings with very different weights at very different 
orders of magnitude. Further research is required to develop techniques to recognize 
such issues, understand the worst-case impact, and mitigate the effect. Since WFSMs are 
generic data structures that are used in many different fields, such research may be 
useful outside of the AIS context.

\section{Conclusion}

AIS were originally invented in the 1990s, a time in which modern ML/AI technology did not 
yet exist, and anomaly detection problems -- especially sequence-based ones -- were hard to 
solve. Nowadays with sequence-based end-to-end learning and transformers, anomaly detection
in sequence data can be approached using state-of-the-art neural network models. Such models
can scale to millions or billions of parameters to learn semantic features of large sequence datasets.
Despite our  contributions in this paper, AIS models are not yet optimized anywhere nearly as well. Crucially, 
such approaches are gradient-based, whereas AIS remains essentially a grandient-free approach.
Even before all this, Stibor wrote already in 2006 that 
(negative selection) ``was thoroughly explored'' and that ``future work in this direction is not meaningful''
\cite{Stibor2006}. Why, then, still bother with AIS? 

We do agree that AIS is unlikely to become a competitive technology for anomaly detection 
in sequence data anytime soon. We disagree, however, that AIS has been ``thoroughly explored''.
For example, our simple experiments with noisy classification problems in Section~\ref{sectionempirical}
have revealed a new fundamental issue with the decades-old negative selection algorithm
\footnote{Technically, we showed these results for positive selection, but the results for
negative selection would be the same (Remark~\ref{remarkbotharethesame}).}: its performance degrades when the input data becomes too large.
This fundamental issue has (to our knowledge) 
never been pointed out before, possibly because without FSM-based AISs, it has not 
been possible to build systems of the scale required to even allow processing of inputs of this size. 
This shows how we need to build such systems at scale to fully understand their properties.

Currently, the most important motivation to carry out this work
is to study the information processing capacity of the immune system itself, which remains
incompletely understood. For example, recent data cast substantial doubt on 
the long-standing immunological theory of negative selection, which the eponymous algorithm
is based on. Repertoire models have been instrumental to place these new findings into context
and to develop a more fine-grained understanding of the function of selection processes in
 the thymus. Again, such models need to be large-scale to be useful, because real immune
systems contain many millions to billions of cells.

Beyond computational immunology, we are intrigued by the fact that the close similarity of AIS
repertoire models and learning classifier systems (LCS) has not been explored further. LCS 
have similar issues as AIS around scale and (doubts about) usefulness, although LCS are typically
used in a reinforcement learning setting where the ability to process very large sets of input
data is not immediately critical. While the AISs we considered in this paper focused mostly 
on the initial stages of repertoire development -- positive and negative selection -- the processes
occurring throughout an individual's lifetime are more akin to reinforcement learning: pathogens
are recognized, eliminated and memorized through an evolution-like process, which in the case
of B~cells also involves mutation of the repertoire sequence and fitness-based selection
(affinity maturation). Therefore, we feel that (repertoire-based) AISs should ultimately 
be seen as a special case of LCS. We hypothesize that the WFSM framework developed in this paper 
should be equally useable to increase the scale of LCS such that richer, more interesting problems
can be studied. We hope that the insights gleaned from such work will allow us to better understand 
parallel distributed information processing systems in Nature, including but not limited to the immune 
system.

\printbibliography

\end{document}